\pdfoutput=1
\documentclass[runningheads]{META/llncs}
\usepackage[table]{xcolor}
\usepackage{url}
\usepackage{hyperref}
\usepackage[normalem]{ulem}
\usepackage{soul}
\usepackage{multirow}
\usepackage{todonotes}
\useunder{\uline}{\ul}{}
\usepackage{xcolor}

\title{Using meaning instead of words to track topics}

\author{Judicael POUMAY\inst{1}\and Ashwin ITTOO\inst{1}}
\authorrunning{J. POUMAY A. ITTOO}
\institute{ULiege/HEC Liege, Rue Louvrex 14, 4000 Liege, Belgium
\email{\{judicael.poumay, ashwin.ittoo\}@uliege.be}}

\date{November 2020}

\begin{document}
\maketitle

\begin{abstract}
The ability to monitor the evolution of topics over time is extremely valuable for businesses.
Currently, all existing topic tracking methods use lexical information by matching word usage.
However, no studies has ever experimented with the use of semantic information for tracking topics.
Hence, we explore a novel semantic-based method using word embeddings.
Our results show that a semantic-based approach to topic tracking is on par with the lexical approach but makes different mistakes. 
This suggest that both methods may complement each other.
\end{abstract}
\section{Introduction} 	
Buried within the voluminous amounts of texts available online are meaningful insights, which could help in supporting business decision-making activities.
Topic modelling methods extracts latent topic in a corpus \cite{LDA,poumay2021htmot} and can be used to discover these insights. Examples of applications include fraud detection \cite{WANG201887}, understanding employee and customer satisfaction \cite{JUNG2019113074,IBRAHIM201937}. Extracted topics can be tracked over time to understand their evolution or discover emerging one. Hence, we focus on this task of topic tracking in which the goal is to link instances of the same topic that have been extracted at different time periods. 

Several methods for tracking topics have been proposed in the past \cite{TTonlineLDA,fan2021clustering,TTscience,TTnews,TTHDP}. These methods use measures such as the JS divergence \cite{TTscience,TTnews,TTHDP} or online topic models  \cite{TTonlineLDA,fan2021clustering} which rely on lexical information to track topic across time. 

However, no studies has ever experimented with using semantic information to track topics over time. Intuitively, semantic based approaches could be promising as they do not rely on simple surface form and can capture concepts such as synonymy. For example, given a topic about "AI", across time we could observe that the term "Machine Learning" has become more popular than "AI". However, a lexical approach to topic tracking would not be able to handle such lexical drift and to relate those words over time. Conversely, such lexical variation would have been captured by a semantic approach. Moreover, topic-word distributions are unstable across multiple runs \cite{agrawal2018wrong}, i.e. the resulting top words of a topic tend to change significantly. This entails that the lexical information we rely upon to track topics is also unstable even if the overall semantic of the topic remains the same. Thus, a semantic-based approach may be more robust.

Hence, our work aims at investigating on the use of semantic information for topic tracking and its comparison against lexical information. Therefore, as our main contribution, we propose a novel semantic topic tracking method known as Semantic Divergence (SD) based on word embeddings. As an ancillary contribution, we study the challenges of topic tracking in the context of hierarchical topic modelling. 

\section{Background and Related work} 

\subsection{Topic Modelling}

LDA \cite{LDA} is the first traditional topic model. At the core of LDA is a Bayesian generative model with two Dirichlet distributions, respectively for the document-topic distributions and for the topic-word distributions. These distributions are learnt and optimized via an inference procedure which enables topics to be extracted. The main weakness of LDA is that it requires the user to specify a predefined number of topics to be extracted. 

More complex topic models have been proposed since  LDA. In particular,  HTMOT \cite{poumay2021htmot} was proposed to simultaneously  model topic hierarchy and temporality. Specifically, HTMOT produces a topic tree in which the depth and the number of sub-topic for each branch is defined dynamically during training. Additionally, HTMOT models the temporality of topics enabling the extraction of topics that are lexically close but temporally distinct.

\subsection{Topic Tracking}
Topic tracking is the task of monitoring the evolution of topics through time. It was initially defined in a pilot study \cite{TDT} in 1998 as the continuous automatic classification of a stream of news stories into known or new topics. 

Currently, two general framework compete for topic tracking.  The first stream is that of online topic models. which incorporate new data incrementally \cite{TTonlineLDA,fan2021clustering}. In \cite{TTonlineLDA}, the authors propose Online-LDA, a version of LDA able to update itself with new documents without having to access to previously processed documents. In practice, Online-LDA assumes that time is divided in slices and at each slice an LDA model is trained using the previous slice as prior. They were able to show that their system can find emerging topics by artificially injecting new topic into the news stream. They performed their experiments on the NIPS and Reuters-21578 datasets. Similarly in \cite{fan2021clustering}, the authors propose a model that can dynamically decide the right number of topics in an online fashion. They performed their experiments on the the 20 Newsgroup and the TDT-2 datasets.

The second stream is concerned with linking topics extracted independently at different time periods \cite{TTscience,TTnews,TTHDP}. In \cite{TTscience}, the authors use about 30,000 abstracts of papers in various journals from 2000 to 2015. They then applied LDA to each year independently and linked topics using the Jensen-Shannon Divergence (JS) to measure their similarity \cite{JS}.  In \cite{TTnews} the authors applied a similar method on news articles. However, they differ in that while \cite{TTscience} simply links topics together,  \cite{TTnews} clusters them. This means that once two topic have been linked they form a cluster and subsequent topics will be compared to the whole cluster and not just the preceding topic.  Finally in \cite{TTHDP}, the authors also proposed a tracking method using the JS divergence applied to scientific papers. However, they do not constraint linkage to a one-to-one mapping which allows for the fusion and splitting of topics. All of the aforementioned paper evaluated their topic tracking method using a qualitative analysis that demonstrated the performance of their technique.

We based our work on that second stream because it allows for  better parallelization as time slices are processed independently.

\section{Methodology}
In this section, we will present our methodology for topic tracking. We will start by describing our corpus and topic extraction method. Next, we will define our SD measure. Finally, we will present the topic tracking algorithm. 

\subsection{Topic extraction}
To perform our experiments, we crawled 10k articles from the Digital Trends \footnote{https://www.digitaltrends.com/} archives from 2019 to 2020. This news website is mainly focused on technological news with topics such as hardware, space exploration and COVID-19. For all articles, we extracted the text, title, category and timestamp. We pre-possessed the corpus according to HTMOT \cite{poumay2021htmot}. 

\begin{figure}
    \centering
    \includegraphics[width=0.8\linewidth]{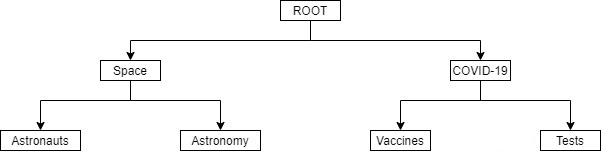}
    \caption{Example of a topic hierarchy}
    \label{fig:extopic}
\end{figure}

To extract topics hierarchies (see figure \ref{fig:extopic}), we used the HTMOT topic model \cite{poumay2021htmot} . The extracted topics are represented by a list of words and a list of entities.

\noindent We follow HTMOT \cite{poumay2021htmot} and only focus on the first and second level of topic extracted. Specifically, the authors observe that deeper topics becomes more esoteric making them harder to understand by annotators representing a general audience. Consequently, this makes it difficult to assess the correctness of tracked topics at deeper levels of the topic tree. 

\subsection{Proposed Semantic Divergence measure}
We will now describe our novel topic tracking method, which departs from the JS divergence traditionally applied in previous studies. We name our method "Semantic divergence" or SD. It uses word embeddings to measure the distance between topics. Each topic will be assigned an embedding as the sum of the embeddings of the top words in that topic weighted by their probability. Then, the distance between two topics is computed as the cosine distance of their respective embedding. We will use FastText as the word embedding. FastText helps  with rare and out of vocabulary words. This is essential considering our pre-processing step includes lemmatization which may produce incorrectly spelled words. Hence the embedding of a topic is defined as follows :

\begin{equation}
    emb(t) = \sum_{(w,p)\in t} p*FastText(w)
\end{equation}

\noindent And the Semantic Divergence between two topics is defined as :
\begin{equation}
SD(t_1,t_2) = cosine(emb(t_1), emb(t_2))    
\end{equation}

\noindent Where $w$ is a word in a topic $t$ and $p$ is the probability of that word.

\subsection{Topic Tracking Algorithm}
Finally, to track topics across time we applied HTMOT on our corpus. For each year (2019 and 2020), we obtained a corresponding topic tree. Then, we computed the distance between every topics across both years using either JS or SD. To do this we used the top 100 words and top 15 entities to represent each topic. Subsequently, we ranked order all computed pairs of topics and then iteratively selected the most similar pairs (lowest SD or JS score) such that each topic is paired only once. Finally, we used a pre-defined threshold to remove pairs with a poor score. 

Note that our approach does not take into account structural information. Indeed, tracking topics in the context of hierarchical topic modelling presents another interesting challenge : there exist many possible resulting trees that are equally correct. In one run, we may extract the topic of space whose sub-topics can be grouped into space exploration and astronomy. Conversely, in another run, we may extract space exploration and astronomy as separate topics with their own sub-topics. Hence, it is difficult to leverage the structural information contained in the topic trees to track topics as it cannot be expected to respect a specific conceptual taxonomy. 
\section{Results : JS vs SD}
In this section, we will discuss how our semantic based method compares with respect to the traditional lexical based method. 

\begin{figure}
    \centering
    \includegraphics[width=0.75\linewidth]{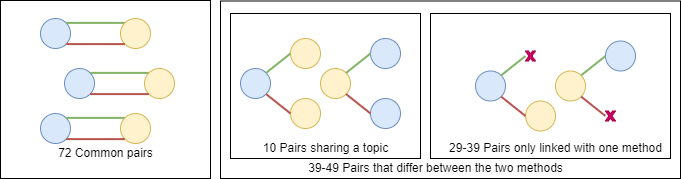}
    \caption{The pairs extracted by both methods can be grouped into three categories. The circle represent topics and their color represent years (2019 blue; 2020 yellow). The link color represent the method used (JS red; SD green). The three categories are : 1) The pairs extracted by both methods (72).  2) The pairs that differ but share a topic (10) E.g. JS extracted the pair 4-D while SD extracted 4-E. 3) The pairs of topics that were only linked with one method (29-39).}
    \label{fig:pairkind}
\end{figure}

First, we studied the overlap between the two methods, i.e. the number of pairs extracted by both.
We discovered that, 111 pairs were extracted with JS with a threshold of $<$0.4, while 121 pairs were extracted with SD with a threshold of $<$0.1. These threshold were set through empirical observation but may depend on the dataset used.
These 111-121 pairs can be grouped into three categories (see figure \ref{fig:pairkind}). 72 pairs were the same between the two methods (60-65\% of the total pairs). For example, topics such as space and video games were easily paired across both years by both methods. This already indicates that our SD method is able to pair topic across time with performance similar to JS. This leaves 39-49 pairs that are different across the two methods (35-40\% of the total pairs) which we can evaluate. Out of those different pairs, we notice that in most cases one method (e.g. SD) would track/link a topic pair across both years, while the other method (e.g. JS) did not as the best possible pair was above the threshold. We are then left with 10 different pairs that can themselves be paired according to which 2019 or 2020 topic they share (see figure \ref{fig:pairkind}).

To compare the performance of the two tracking methods, we decided to use a survey comparing these 10 pairs of topics extracted by both JS and SD. Precisely, for each question, given an initial topic, annotators were shown the JS and SD pairing and asked which is better. Additionally, we also asked annotators to provide a confidence score on a scale from 1 to 5. In total, we received 38 answers coming from a small online community focused on answering surveys\footnote{\url{https://www.reddit.com/r/SampleSize/}}. The survey can be found on github
\footnote{\url{https://github.com/JudicaelPoumay/TopicTrackingPaper}}.

Looking at the survey results (table \ref{surveyres}), it can be seen that SD slightly outperforms JS with 54\% of annotators preferring the former to the latter. Moreover, we also note that the annotators were confident in their evaluation, with an average confidence score of 3.3. Interestingly, there is a lot of variability in the answers. Some topics were clearly better paired with one method or the other (Q3 and Q5) while for others, it wasn't as clear (Q1, Q2 and Q4).

\begin{table}[h]
\centering
\caption{The "chose SD" column corresponds to the \% of annotators that chose the SD pair as the best pair.}
\begin{tabular}{l|ll}
\rowcolor[HTML]{C0C0C0} 
Questions & \multicolumn{1}{l|}{\cellcolor[HTML]{C0C0C0} Chose SD} & Confidence level \\ \hline
Q1 & 42.1\% (22) & 3.2 \\
Q2 & 63.2\% (14) & 2.6 \\
Q3 & 21.1\% (30) & 3.7 \\
Q4 & 65.8\% (13) & 3.5 \\
Q5 & 78.9\% (8)  & 3.5 \\
Average & 54\%  & 3.3
\end{tabular}
\label{surveyres}
\end{table}

\begin{figure}
    \centering
    \includegraphics[width=0.65\textwidth]{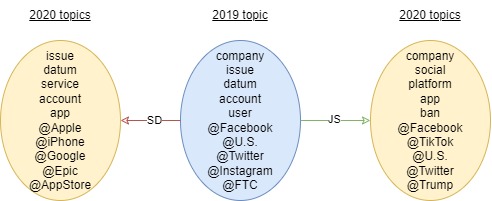}
    \caption{A first example of different pairing  between SD and JS on the same 2019 topic.}
\label{tab:exJSSD1}
\end{figure}

\noindent For example, figure \ref{tab:exJSSD1} corresponds to Q1. It shows how a 2019 topic has been paired with 2020 topics using JS and SD. 
First, we can notice that the distance recorded between the pairs is close to the threshold for both methods.
Specifically, 0.29 for the JS pair and 0.09 for the SD pair (threshold = 0.4 for JS and 0.1 for SD).
This makes sense as good pairs (pairs with low  JS/SD values) are extracted by both methods.
Second, the 2019 topic is about social media data security.
Whereas the chosen 2020 topic is about :
\begin{itemize}
\item Social media when paired with JS.
\item Data security when paired with SD.
\end{itemize}
Hence, both pairing seems suitable, which could explain the indecisiveness of annotators.
Specifically, 16 of them decided the SD pairing was better whereas 22 of them decided the JS pairing was better.
Their confidence level for this question was 3.2 out of 5.

\begin{figure}
    \centering
    \includegraphics[width=0.65\linewidth]{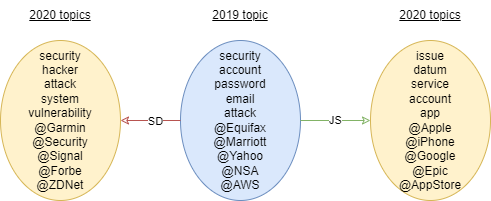}
    \caption{A second example of different pairing  between SD and JS on the same 2019 topic.}
\label{tab:exJSSD2}
\end{figure}

Similarly, figure \ref{tab:exJSSD2} corresponds to Q5 and shows how another 2019 topic has been paired based on the two methods. 
Here, the 2019 topic is about web security.
Whereas the chosen 2020 topic is about :
\begin{itemize}
\item Data security when paired with JS.
\item Web security topic when paired with SD. 
\end{itemize}
Moreover, the topic chosen by SD is a sub-topic of the topic chosen by JS which demonstrates the difficulty in topic tracking in a hierarchical setting.
Indeed, it can be difficult to differentiate a topic from its sub-topic, especially if that sub-topic dominates the others as parent topics are the sum of their sub-topics.
In this case, annotators agreed more and 30 out of 38 decided the SD pair was better.
Their confidence level for this question was 3.5 out of 5.

\noindent Hence, we argue that JS and SD are two fundamentally different approaches and that both have their advantages. 
JS is lexically driven and may work best for linking topics which tend to have a stable and precise vocabulary such as in legal documents.
On the other hand, SD is driven by semantics and may be more appropriate for linking topics that have a greater lexical variability. Greater lexical variability may be the result of lexical drift over time as terms change in popularity or informal texts which do not use a standard vocabulary such as tweets. 
Hence, we believe that SD not only competes but complements JS for topic tracking. 
\section{Conclusion} 	
In this paper, we presented a novel semantic-based topic tracking method (SD). We showed that its performance was comparable to that of the state of the art method (JS), which is lexically-based. This validates our hypothesis that semantic information is valuable for tracking topics. 

Moreover, we have discussed the challenges associated with tracking topics in a topic hierarchy. First, topics and their sub-topic can be difficult to differentiate, which makes topic tracking more challenging. Second, deeper topics are more esoteric and consequently it is harder to assess the quality of their tracking. Finally, topic hierarchy may have many equally correct arrangements which makes it difficult to leverage structural information for topic tracking.

We believe that our work would benefit future studies investigating hybrid methods for topic tracking, such as by integrating lexical and semantic information.

\bibliographystyle{META/splncs04}
\bibliography{META/anthology, META/bib}

\begin{thebibliography}{10}
\providecommand{\url}[1]{\texttt{#1}}
\providecommand{\urlprefix}{URL }
\providecommand{\doi}[1]{https://doi.org/#1}

\bibitem{agrawal2018wrong}
Agrawal, A., Fu, W., Menzies, T.: What is wrong with topic modeling? and how to
  fix it using search-based software engineering. Information and Software
  Technology  \textbf{98},  74--88 (2018)

\bibitem{TDT}
Allan, J., Carbonell, J.G., Doddington, G., Yamron, J., Yang, Y.: Topic
  detection and tracking pilot study final report  (1998)

\bibitem{TTonlineLDA}
AlSumait, L., Barbar{\'a}, D., Domeniconi, C.: On-line lda: Adaptive topic
  models for mining text streams with applications to topic detection and
  tracking. In: 2008 eighth IEEE international conference on data mining. pp.
  3--12. IEEE (2008)

\bibitem{LDA}
Blei, D.M., Ng, A.Y., Jordan, M.I.: Latent dirichlet allocation. J. Mach.
  Learn. Res.  \textbf{3}(null),  993–1022 (Mar 2003)

\bibitem{JS}
Dagan, I., Lee, L., Pereira, F.: Similarity-based methods for word sense
  disambiguation. In: 35th Annual Meeting of the Association for Computational
  Linguistics and 8th Conference of the {E}uropean Chapter of the Association
  for Computational Linguistics. pp. 56--63. Association for Computational
  Linguistics, Madrid, Spain (Jul 1997). \doi{10.3115/976909.979625},
  \url{https://aclanthology.org/P97-1008}

\bibitem{fan2021clustering}
Fan, W., Guo, Z., Bouguila, N., Hou, W.: Clustering-based online news topic
  detection and tracking through hierarchical bayesian nonparametric models.
  In: Proceedings of the 44th International ACM SIGIR Conference on Research
  and Development in Information Retrieval. pp. 2126--2130 (2021)

\bibitem{IBRAHIM201937}
Ibrahim, N.F., Wang, X.: A text analytics approach for online retailing service
  improvement: Evidence from twitter. Decision Support Systems  \textbf{121},
  37--50 (2019). \doi{https://doi.org/10.1016/j.dss.2019.03.002},
  \url{https://www.sciencedirect.com/science/article/pii/S0167923619300405}

\bibitem{JUNG2019113074}
Jung, Y., Suh, Y.: Mining the voice of employees: A text mining approach to
  identifying and analyzing job satisfaction factors from online employee
  reviews. Decision Support Systems  \textbf{123},  113074 (2019).
  \doi{https://doi.org/10.1016/j.dss.2019.113074},
  \url{https://www.sciencedirect.com/science/article/pii/S0167923619301034}

\bibitem{TTHDP}
Liu, H., Chen, Z., Tang, J., Zhou, Y., Liu, S.: Mapping the technology
  evolution path: a novel model for dynamic topic detection and tracking.
  Scientometrics  \textbf{125}(3),  2043--2090 (2020)

\bibitem{poumay2021htmot}
Poumay, J., Ittoo, A.: {HTMOT} : {H}ierarchical {T}opic {M}odelling {O}ver
  {T}ime (2021). \doi{10.48550/arXiv.2112.03104}

\bibitem{WANG201887}
Wang, Y., Xu, W.: Leveraging deep learning with lda-based text analytics to
  detect automobile insurance fraud. Decision Support Systems  \textbf{105},
  87--95 (2018). \doi{https://doi.org/10.1016/j.dss.2017.11.001},
  \url{https://www.sciencedirect.com/science/article/pii/S0167923617302130}

\bibitem{TTnews}
Xu, G., Meng, Y., Chen, Z., Qiu, X., Wang, C., Yao, H.: Research on topic
  detection and tracking for online news texts. IEEE access  \textbf{7},
  58407--58418 (2019)

\bibitem{TTscience}
Zhu, M., Zhang, X., Wang, H.: A lda based model for topic evolution: Evidence
  from information science journals. In: Proceedings of the 2016 International
  Conference on Modeling, Simulation and Optimization Technologies and
  Applications (MSOTA 2016). pp. 49--54 (2016)

\end{thebibliography}

\end{document}